\documentclass[sigconf]{acmart}

\usepackage{booktabs} 
\usepackage{float}
\usepackage[utf8]{inputenc}
\usepackage{graphicx}
\usepackage{placeins}
\usepackage{enumitem} 

\setcopyright{none}

\acmDOI{}

\acmISBN{}

\acmConference[SIGIR'18]{ACM SIGIR Conference}{July 2018}{Ann Arbor, Michigan, USA} 

\acmArticle{}
\acmPrice{}

\setcopyright{rightsretained}
\acmConference[SIGIR 2018 eCom]{ACM SIGIR Workshop on eCommerce}{July 2018}{Ann Arbor, Michigan, USA} 
\acmYear{2018}
\copyrightyear{2018}

\begin{document}
\title{Visualizing and Understanding Deep Neural Networks in CTR Prediction}

\author{Lin Guo}
\affiliation{%
  \institution{Alibaba Group}
}

\author{Hui Ye}
\affiliation{%
  \institution{Alibaba Group}
}

\author{Wenbo Su}
\affiliation{%
  \institution{Alibaba Group}
}

\author{Henhuan Liu}
\affiliation{%
  \institution{Alibaba Group}
}

\author{Kai Sun}
\affiliation{%
  \institution{Alibaba Group}
}

\author{Hang Xiang}
\affiliation{%
  \institution{Alibaba Group}
}


\begin{abstract}
Although deep learning techniques have been successfully applied to many tasks, interpreting deep neural network models is still a big challenge to us. Recently, many works have been done on visualizing and analyzing the mechanism of deep neural networks in the areas of image processing and natural language processing. In this paper, we present our approaches to visualize and understand deep neural networks for a very important commercial task---CTR (Click-through rate) prediction. We conduct experiments on the productive data from our online advertising system with daily varying distribution. To understand the mechanism and the performance of the model, we inspect the model's inner status at neuron level. Also, a probe approach is implemented to measure the layer-wise performance of the model. Moreover, to measure the influence from the input features, we calculate saliency scores based on the back-propagated gradients. Practical applications are also discussed, for example, in understanding, monitoring, diagnosing and refining models and algorithms.

\end{abstract}

%
%



\maketitle

\section{Introduction}
Click-through rate (CTR) prediction plays a crucial role in computational advertising. In the common cost-per-click advertising system, advertisements are ranked by the product of the bid price and the predicted CTR when bidding for impression opportunities. Therefore, the revenue of the multi-billion business heavily relies on the performance of the CTR prediction model.

Deep learning techniques have been successfully applied to CTR prediction tasks ~\cite{deepcross,widedeep,youtube}. Deep neural networks (DNNs), composed of stacked layers of neurons, have the capability to extract the nonlinear patterns from features and thus reduce the burden of nontrivial feature engineering. However, the working mechanisms of deep learning models are still not well understood. The lack of interpretability becomes an obstacle for deep learning, and raises concerns on the reliability of deep learning applications, especially for critical industrial implementations.

Many recent progresses have been made in visualizing and interpolating deep learning models for image processing ~\cite{szegedy2013intriguing,mahendran2016visualizing,zeiler2014visualizing,koh2017understanding,rauber2017visualizing,pei2017deepxplore} and natural language processing ~\cite{karpathy2015visualizing,jiewei2016visualizing,Tang2017Memory,arras2017explaining,bahdanau2014neural}. In this paper, we present a series of approaches to visualize and analyze a simple DNN model for CTR prediction on the productive data from our search advertising platform. The model's performance decay is investigated over datasets with daily varying distribution, and the distributions of the output scores are also compared for different training stages. We inspect the model's inner status down to neuron level. We study the statistical properties of the neurons' statuses for the hidden layers, and investigate the high-level representations learned by the model through t-SNE projection ~\cite{rauber2017visualizing,maaten2008visualizing}. A probe method ~\cite{alain2016understanding} is applied to dissect model's performance layer by layer for different datasets. Moreover, to measure the influence of the input features, we calculate saliency scores for the feature groups based on back-propagated gradients. 

Beyond the classic model evaluation metrics ~\cite{Graepel:2010:WBC:3104322.3104326,goodfellow2016deep}, we open up the "black box" and inspect the DNN model from the output to the input end. Understanding the model's  mechanism can help us not only design and diagnose models, but also monitor the algorithmic advertising system for daily production.

\section{Experimental Setting}
\subsection{Datasets}
\label{sec:expdata}
We perform experiments on the productive CTR prediction data from the search advertising platform of our company. Started from a typical Wednesday, our data are collected over eight consecutive days. The training set is sampled from day one. To investigate decay of the model's performance, we evaluate the model on a daily basis from day one to day eight. The eight test sets are, in turn, denoted by test$1$, test$2$, ..., test$8$. Each dataset contains about $150$ million instances which are randomly sampled from the ad impression logs of the corresponding day. Note that there are no overlap between test$1$ and the training set. The setup of datasets simulates the real world environment for the CTR prediction task, i.e., the model is trained with historical data and deployed to serve the future online traffic, where the data distribution varies and differs with the training data by nature.

Our data contains $34$ groups of sparse categorical features (around $100$ million binary features in total), e.g., user id, user's city, user's gender, user's age level, query id, query words, shop id, ad's category, etc.. Note that there are no combinational features in this study.

\subsection{Model setting}
\label{sec:expmodel}

The DNN model contains four fully-connected hidden layers. From layer $1$ (closest to input) to layer $4$ (right before output layer), the layer's width is set to $256$, $128$, $64$ and $32$ neurons. The formulation for the output vector of $k$th hidden layer, denoted by $\mathbf{h}_{k} $, can be written as: 
\begin{eqnarray}
\label{eqn:1.1}
\mathbf{h}_{k} = ReLU( \mathbf{W}_{k}\mathbf{h}_{k-1}+ \mathbf{b}_{k}  ),
\end{eqnarray}
Where  $\mathbf{W}_{k}$ is the weight tensor of all the connections from the neurons of layer $(k-1)$, $\mathbf{b}_{k}$ represents the bias term and ReLU (rectifier linear unit) function is used as the activation function. The output layer uses a sigmoid function to map the output to a float number between $0$ and $1$ as the predicted probability of click:

\begin{eqnarray}
\label{eqn:1.2}
Pctr = Sigmoid( \mathbf{W}_{5}\mathbf{h}_{4}+ \mathbf{b}_{5}  ).
\end{eqnarray}
For the training process, $Pctr$ is compared against the ground truth label and cross entropy is calculated as the loss function. For each input instance, the sparse feature ids are embedded into 8-dimensional float vectors ~\cite{widedeep,youtube,deepcross}. For feature groups containing multiple feature ids per instance, e.g., query words, sum pooling operations are applied to enforce each feature group to produce an 8-dimensional embedding vector. The embedding outputs are concatenated into a 272-dimensional vector, denoted by $\mathbf{h}_{0}$, as the input to layer $1$. The embedding vectors are trained jointed with the other parts of the model.

The experiments are run on distributed TensorFlow~\cite{abadi2016tensorflow} released by Google. The model is trained by Adagrad optimizer ~\cite{duchi2011adaptive} with learning rate $=0.005$, initial accumulator value $=0.0001$ and mini-batch size $=1000$. Glorot and Bengio's method ~\cite{Glorot2010Understanding} is used for initialization. We visualize the model's inner status by dynamically dumping the processing data based on model graph.

\section{Results}
\label{sec:res}

\subsection{AUC and Prediction Score}
\label{sec:resauc}

\begin{figure}[htb!]
\includegraphics[width=3.3in]{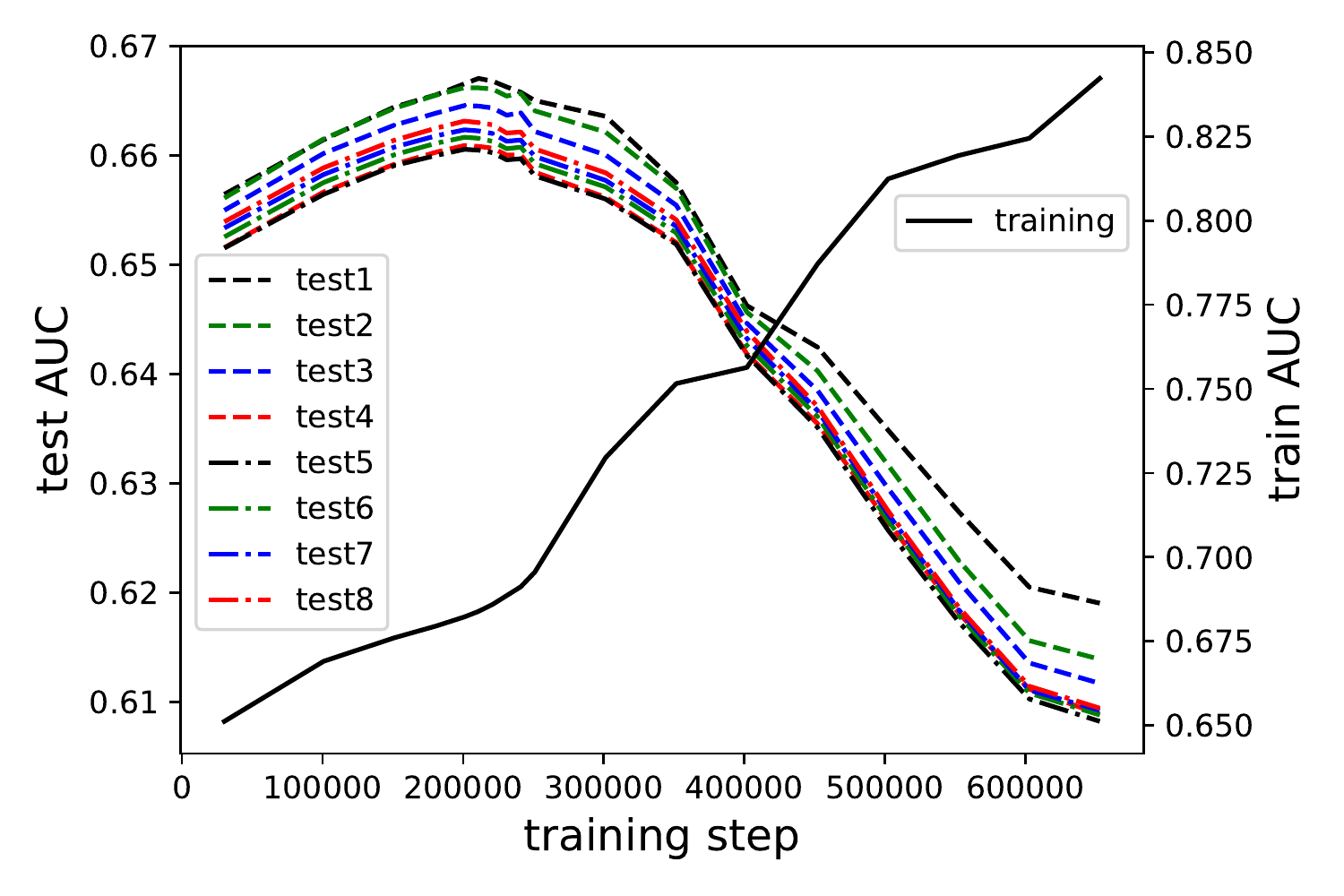}
\caption{AUC score as a function of training step for training and test sets.}
\label{fig:auc}
\end{figure}

To measure the performance of model, we employ AUC (area under curve of the receiver operating characteristic plot) as the key metric. AUC is a widely used measure for evaluating the CTR performance ~\cite{Graepel:2010:WBC:3104322.3104326}.

In Fig. \ref{fig:auc},  we present the evolution of the model's AUC  as a function of the training steps for training and test sets. With the training going on, the train AUC keeps growing, while all the test AUCs follow a same pattern --- first rises  and then decreases due to overfitting. 
The model generalizes best at step $210000$. Comparing the eight test AUCs for the same time step, the model's performance decay can be disclosed as a function of dataset. The test AUC score decreases monotonically from day one to day five. As expected, this is because the distribution of the test data differs with the training set, and the difference grows day by day. After that, AUC upswings for the last three days and surpasses day four. This is in accordance with a characteristic of our business scene --- although the data varies from day to day, the users' behaviors on our website have weekly periodic patterns. This non-monotonic change of AUC is evident for the regime from under-fitting to weak overfitting (before step $\sim 400000$). At larger training steps, overfitting becomes severe and the model performs same bad for the last five days.

\begin{figure}[htb!]
\includegraphics[width=3.5in]{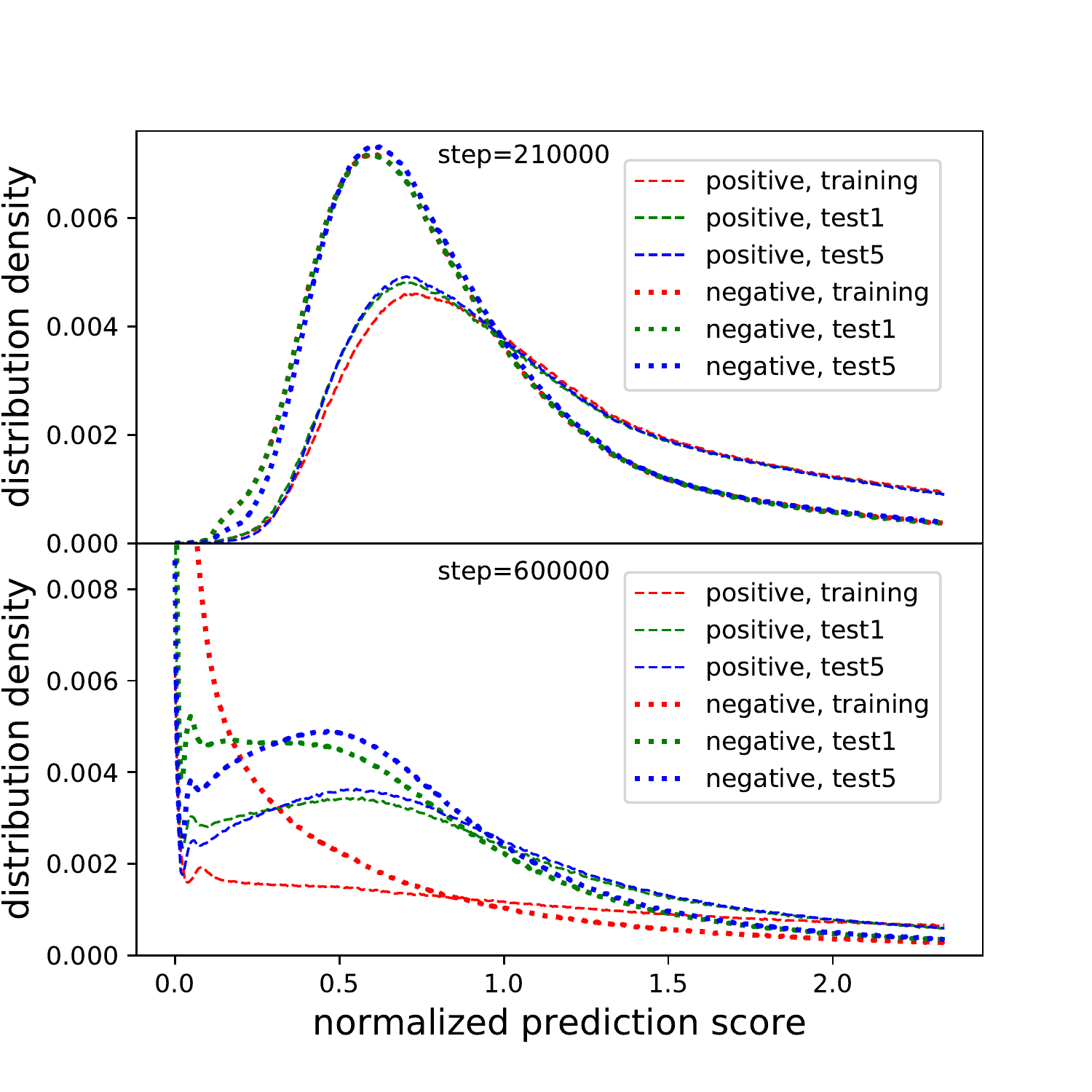}
\caption{Distribution of predicted CTR for models at training step 210000 and step 600000. The X-axis denotes the predicted CTR normalized by the average click ratio of the training set.}
\label{fig:scoredist}
\end{figure}
\begin{figure*}[htb]
	\includegraphics[width=5.5in]{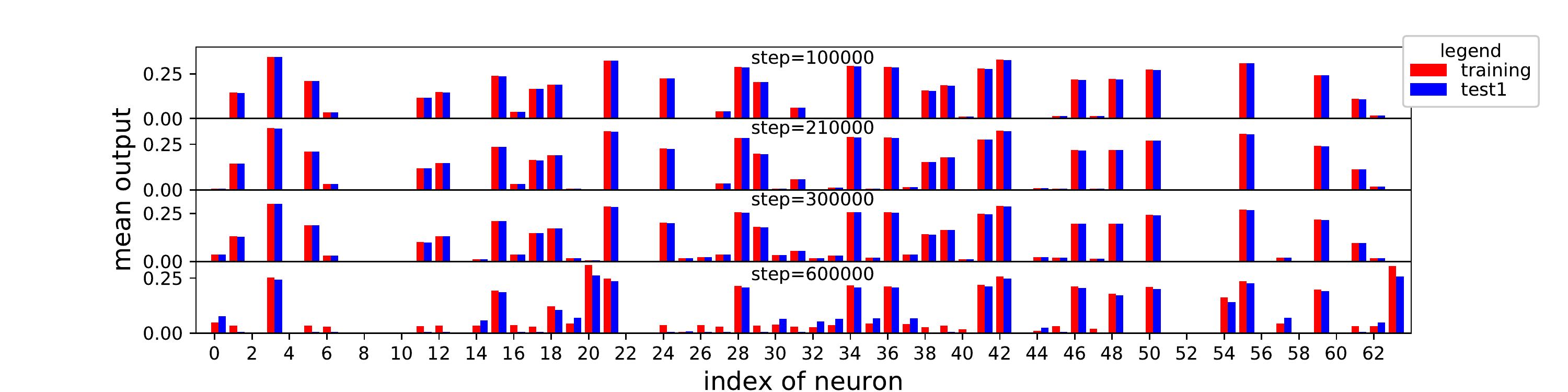}
	\caption{Mean outputs of the neurons in layer $3$ for training and test1 sets, for training step 100000, 210000, 300000 and 600000. Each bar represents a neuron.}
	\label{fig:statsmeanlayer3}
\end{figure*}

\begin{figure*}[htb]
	\includegraphics[width=5.5in]{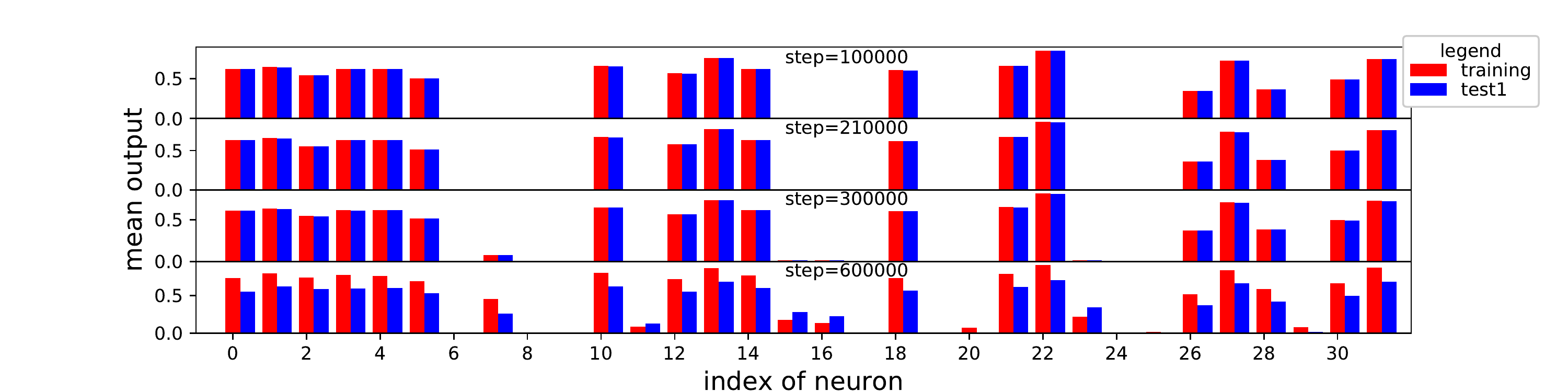}
	\caption{Mean outputs of the neurons in layer $4$ for training and test1 sets, for training step 100000, 210000, 300000 and 600000. Each bar represents a neuron.}
	\label{fig:statsmeanlayer4}
\end{figure*}

\begin{figure*}[htb]
\includegraphics[ width=5.5in]{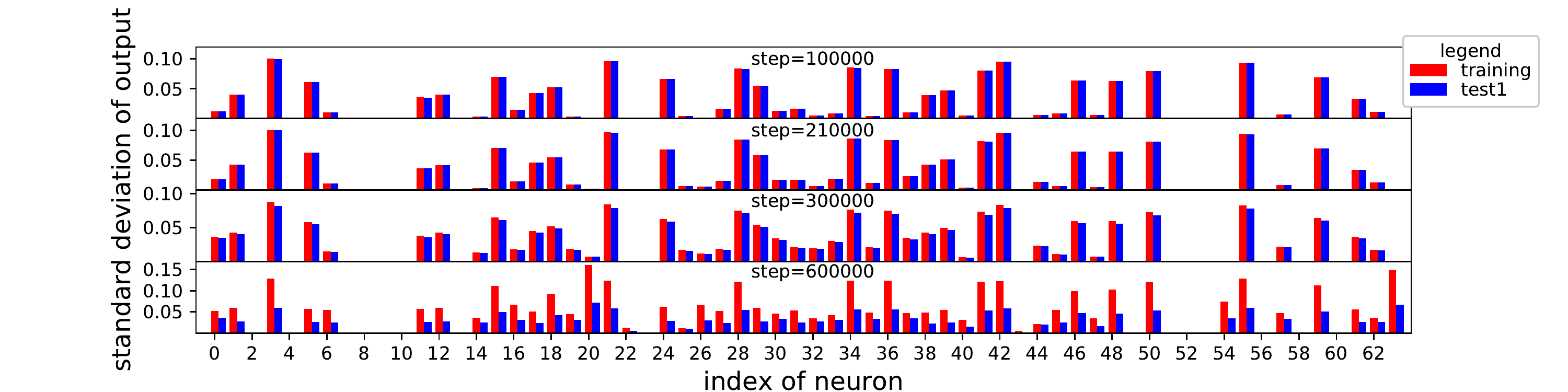}
\caption{Standard deviations of the outputs of the neurons in layer $3$ for training and test1 sets, for training step 100000, 210000, 300000 and 600000. Each bar represents a neuron.}
\label{fig:statsstdlayer3}
\end{figure*}

\begin{figure*}[htb]
\includegraphics[width=5.5in]{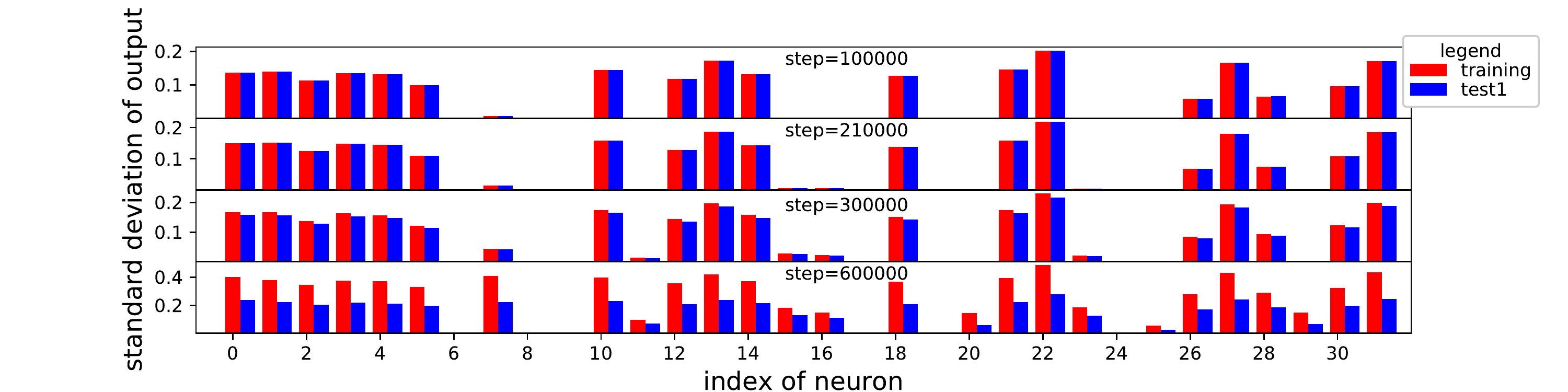}
\caption{Standard deviations of the outputs of the neurons in layer $4$ for training and test1 sets, for training step 100000, 210000, 300000 and 600000. Each bar represents a neuron.}
\label{fig:statsstdlayer4}
\end{figure*}

Fig. \ref{fig:scoredist} provides insights into the distribution of predicted CTR score for training, test1 and test5 sets. At training step $210000$, the AUC decay from training set to test1 is mainly because the CTR of the positive (clicked) samples in test1 are more under-predicted by the model. The further decay from test1 to test5 is mainly due to that the negative (non-clicked) samples in test5 tend to be predicted with higher CTRs (the train and test1 curves overlap for the negative samples and can hardly be distinguished by eye). For training step $600000$, the model overfits the training data such that it aggressively predicts the CTR towards zero for both clicked and non-clicked samples. This is attributed to the high skewness of the data. The proportion of clicked samples is lower than $10\%$, so under-predicting the CTR for all samples may still reduce loss in training. This shape of distribution changes significantly as the data become different, the scores move rightwards and the distribution becomes blurred.

\subsection{Neuron Status}
\label{sec:reslayer}

In this subsection, we investigate the statistics of the neurons' statuses for different training stages and datasets. These statistical properties depict the model's representation of the input data, and can help us to interpret the model's performance and working mechanism.

\begin{figure}[htb!]
\includegraphics[width=2.9in]{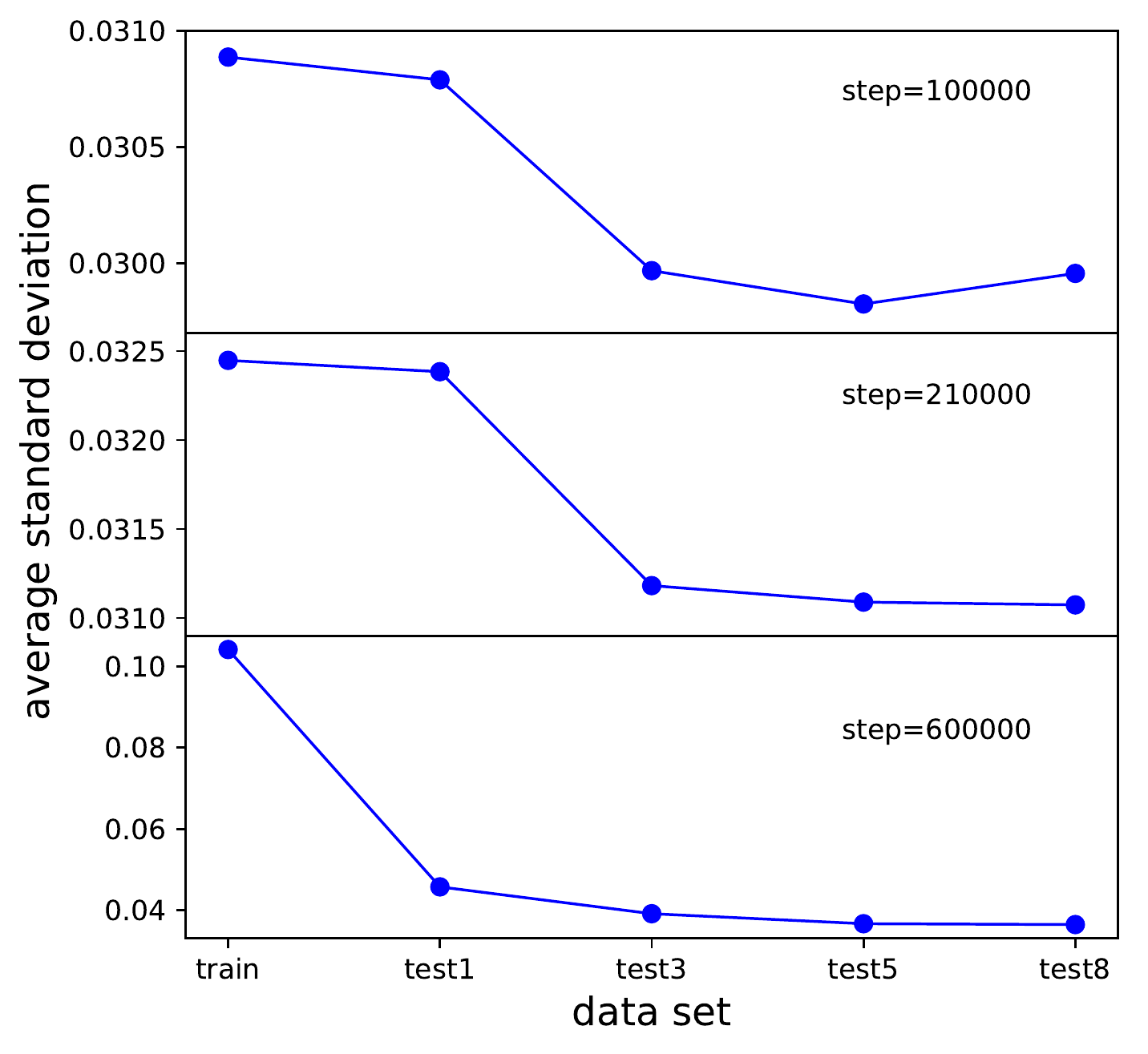}
\caption{Average standard deviation of neurons' outputs for layer $3$ as a function of dataset, for training step 100000,  210000 and 600000. The output's  standard deviation is first calculated for each neuron, and then averaged over all the $64$ neurons of layer $3$. }
\label{fig:stdlayer3}
\end{figure}

The mean outputs of the neurons within layer $3$ and $4$ are illustrated in Figs. \ref{fig:statsmeanlayer3} and \ref{fig:statsmeanlayer4}, respectively. Correspondingly, the standard deviation of the neurons' outputs are plotted in Figs. \ref{fig:statsstdlayer3} and \ref{fig:statsstdlayer4}. For step $100000$ and $210000$, the results are quite close between the underfitting and well-fitting stages. About a quarter of the neurons are barely activated.  Significant changes are observed for the overfitting regime (step > $300000$). More neurons become activated. Also, the difference between the training and test sets grows with the degree of overfitting, especially in the standard deviation (Figs. \ref{fig:statsstdlayer3} and \ref{fig:statsstdlayer4}). The higher standard deviation on the training set indicates that the neurons become over sensitive to the input of the training data. Fig. \ref{fig:stdlayer3} presents the variation of the standard deviation averaged over all the $64$ neurons of layer $3$ as a function of dataset. For all the three different training stages, the trend of the average standard deviation correlates with the model's AUC score (Fig. \ref{fig:auc}).

\begin{figure}[htb!]
\centering
\includegraphics[width=3in]{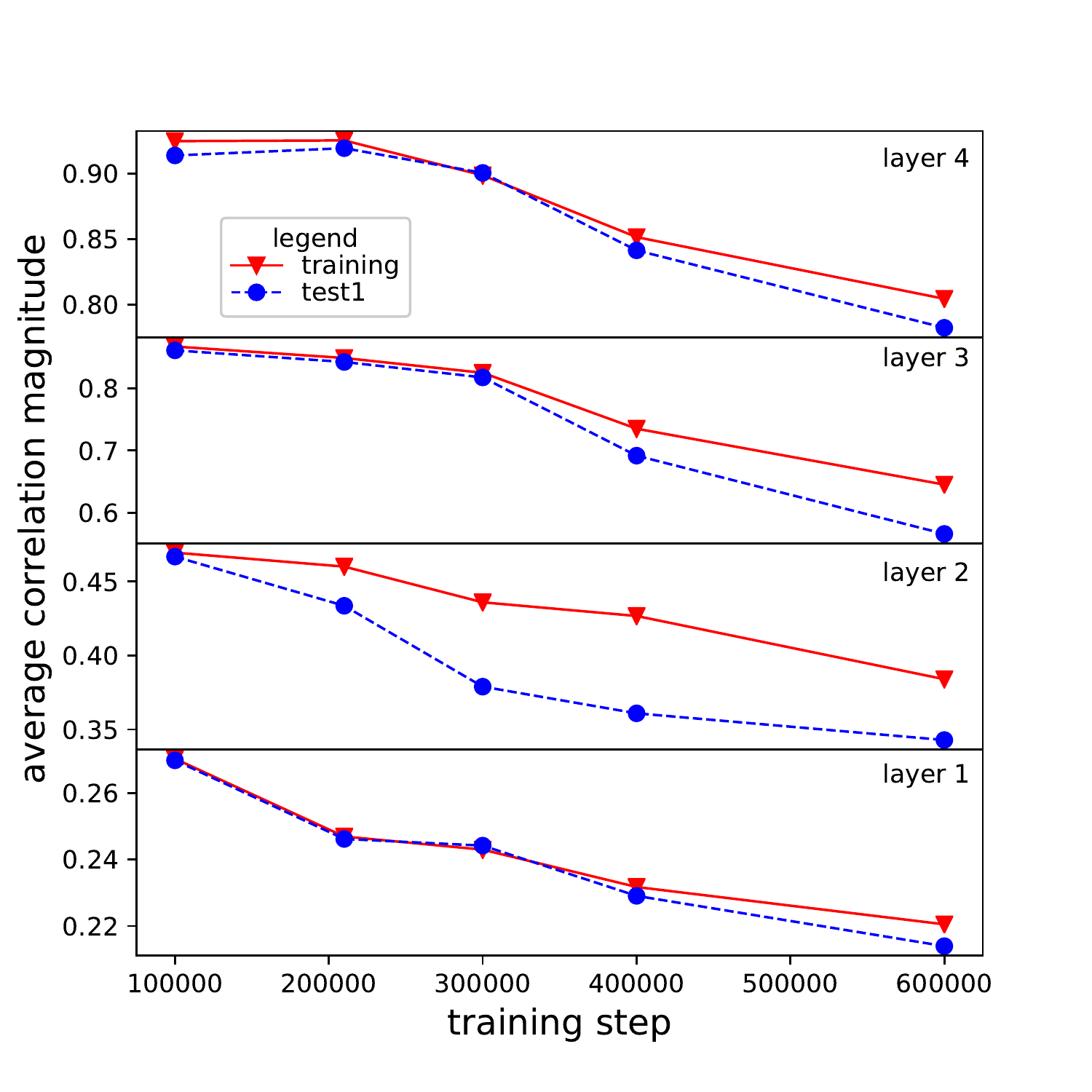}
\caption{Average magnitude of the correlations among the neurons for each hidden layer. The evolution as a function of training step is plotted for training and test1 set.}
\label{fig:corrstep}
\end{figure}

To gain more knowledge about the collaborative patterns of neurons inside the model ~\cite{szegedy2013intriguing,rauber2017visualizing}, for each layer, we calculate the correlations among the neurons. Neurons' statuses before activation are used. We measure the average degree of neurons' correlations  by averaging the absolute value of all the correlation coefficients for each layer. The average strength of correlations is plotted as a function of training step in Fig. \ref{fig:corrstep}. The degree of correlation climbs up with the height of layer. This indicates that the DNN model is refining the input information through the successive layers ~\cite{Tishby2015deep,shwartz2017opening,michael2018on}. Only very limited portion of the input information can be transfered to the output layer.  

After step $210000$, the neurons' correlation deceases monotonically with training step for all hidden layers. Recalling the enhanced neuron activation observed for this overfitting regime (Figs. \ref{fig:statsmeanlayer3} and \ref{fig:statsmeanlayer4}), we can interpret that the model starts to explore more predictive patterns from the input information. However,  the deceasing test AUC (Fig. \ref{fig:auc}) reveals that the boosted representation of the input from training data can not be well generalized to predict the test data.

\begin{figure*}[htb!]
\centering
\includegraphics[width=5in]{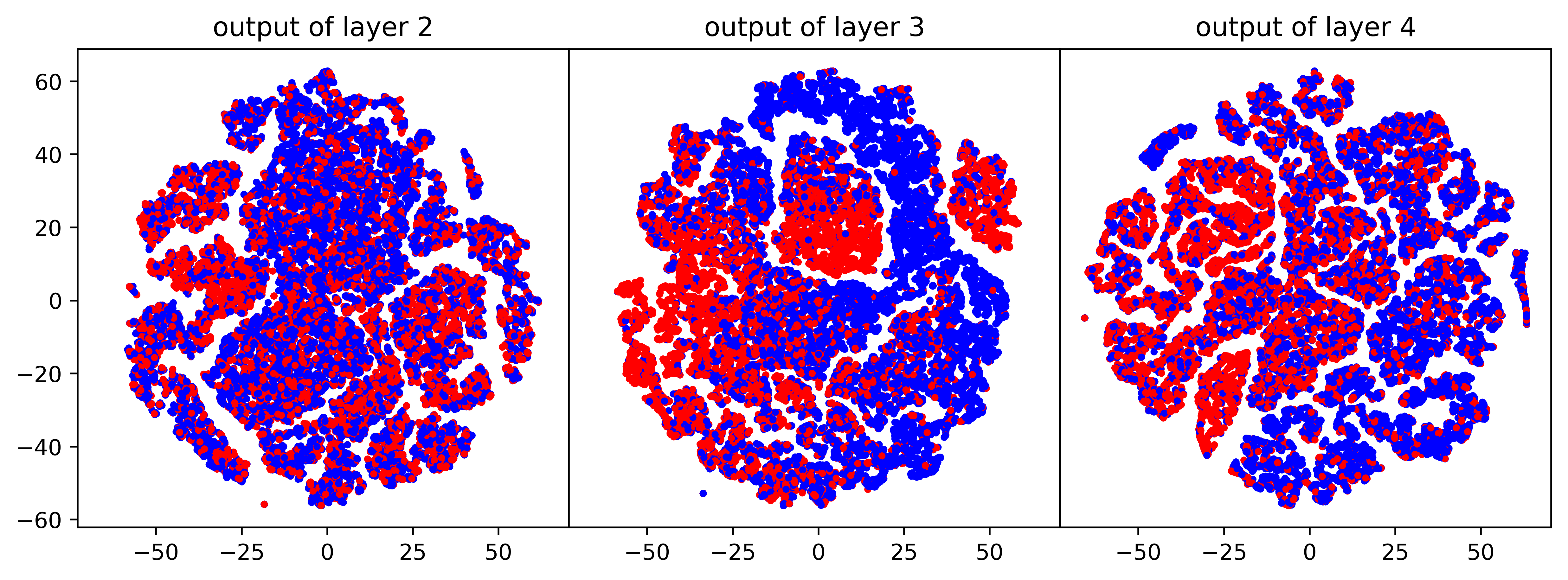}
\caption{Visualization of the output vectors for layer $2$, $3$ and $4$ using t-SNE method, at training step $210000$. Clicked and non-clicked samples are represented by red and blue points, respectively }
\label{fig:tsne}
\end{figure*}

In order to inspect the spacial structure of the high-level representations for the input data, we project the neurons' output vectors to 2-dimensional space using t-SNE method ~\cite{rauber2017visualizing,maaten2008visualizing}. The t-SNE projection is able to preserve neighborhoods and clusters of the data points in the original representation space. In Fig. \ref{fig:tsne}, we illustrate the projection results for layer $2$, $3$ and $4$  at training step $210000$. The presented $10000$ clicked and $10000$ non-clicked instances are randomly selected from the training set. 

For layer $3$ (the center plot in Fig. \ref{fig:tsne}), we can clearly see the regions with concentrated clicked points. We find that the training process enhances the concentration of clicked points for the training set, indicating that the model learns more discriminative representation for the training data.  For the test datasets, we observe that the concentrated distribution disappears when overfitting happens. Unlike the case of image classification in Ref. ~\cite{rauber2017visualizing}, no class separation is observed even at severely overfitting stage. This is mainly due to the highly noisy and skewed data for the CTR prediction task. 

Comparing with the left plot in Fig. \ref{fig:tsne}, the concentration of clicked points of layer $2$ is obviously worse than layer $3$. This agrees with the assumption that for a properly trained DNN model, the discriminative quality of the hidden layer's output increases with the height of the layer~\cite{rauber2017visualizing,alain2016understanding,bengio2009learning}. However, as revealed in the right plot of Fig. \ref{fig:tsne}, the clicked points for  layer $4$ show no improvement in the degree of concentration and look even slightly more scattered. Recalling the very strong correlations among the neurons in layer $4$ (Fig. \ref{fig:corrstep}), one may doubt whether the output of layer $4$ is more predictive than layer $3$. This issue will be further discussed in the following subsections.

\subsection{Probe Evaluations}
\label{sec:resprobe}

\begin{figure}[htb!]
\includegraphics[width=3in]{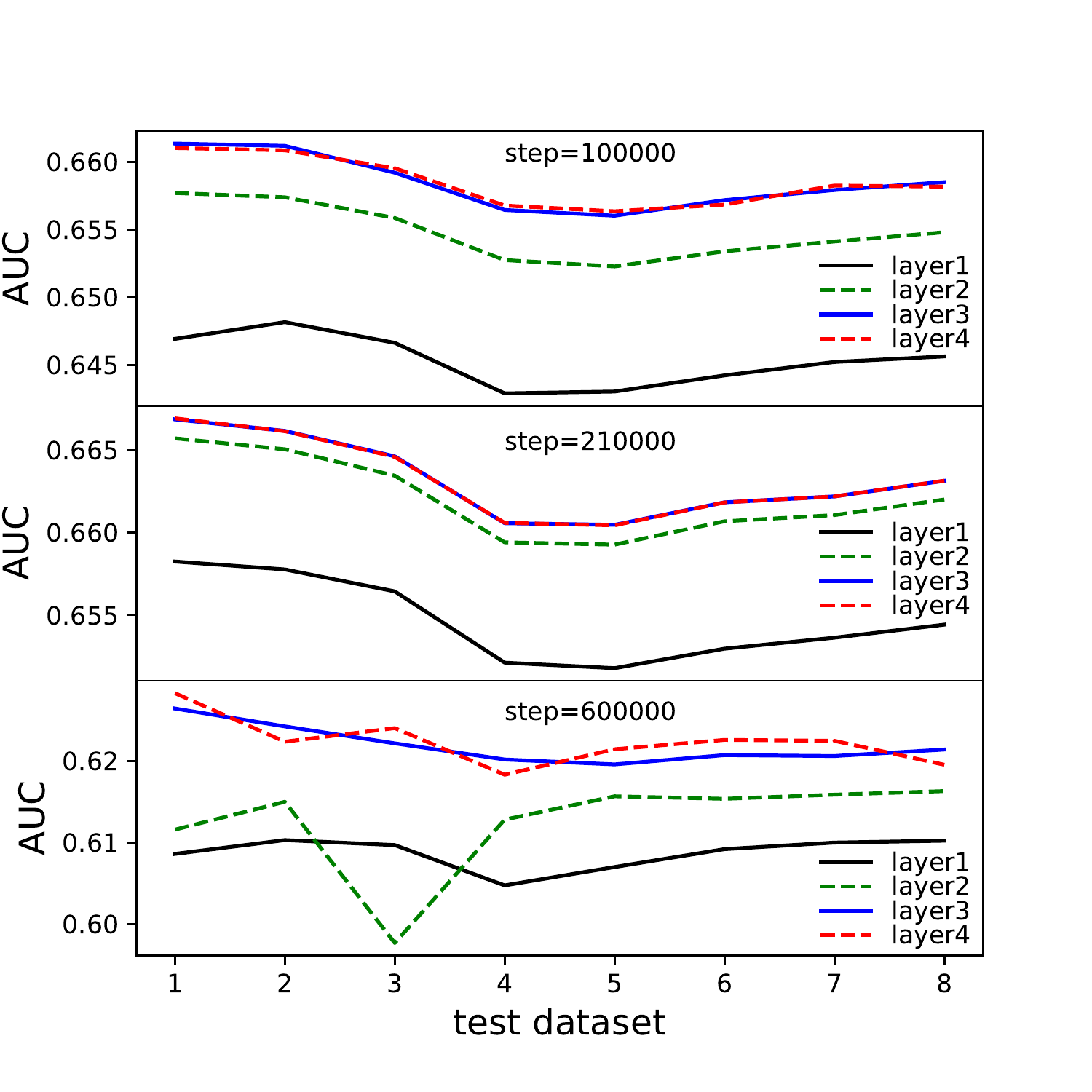}
\caption{Test AUC scores of the probe LR models as a function of test dataset for three training steps: 100000, 210000 and 600000.}
\label{fig:probeauc}
\end{figure}

To investigate the effectiveness of the hidden layers, we implement Alain $\&$ Bengio's  probe approach ~\cite{alain2016understanding}. DNN model is expected to mining for predictive patterns from input features through layers of transformations, and then feed the extracted information into the simple linear classifier at the output end. For each layer,  we use the layer's output vector as input features to train a LR (Logistic Regression) model to predict CTR. The LR model serves as a probe to evaluate the usefulness the hidden layer. A higher performance of the LR probe implies that the transformation of this layer makes information more predictive, and thus benefits the performance of the whole DNN model.

The LR models are trained on the data of the training set until convergence, with the DNN model fixed, and then the performances are evaluated on the tests sets. As shown in Fig. \ref{fig:probeauc}, for training step $210000$,  the performance increases from layer $1$ to layer $3$, indicating that these layers do transform input information to be more predictive. The probe's performance for layer $4$ is the same as layer $3$, indicating that layer $4$ is not as useful as the previous three layers. This is consistent with the observations in the last subsection.

The change of AUC along each curve (in Fig. \ref{fig:probeauc}) illustrates how the hidden layer reacts to the varying data distribution.  At training step $210000$ where the DNN model generalizes best, the effectiveness of all the layers varies as a function of dataset in the same pattern with the DNN model. In contrast, for training step $100000$, where the DNN model is underfitting,  layer $1$ behaves differently with the other layers. Moreover, for  step $600000$, the DNN model overfits the  training data such that the learned information transformations begin to fail for test data. Therefore, the performance of probes is very low and fluctuates significantly.

\subsection{Feature Group Saliency}
\label{sec:resgrad}

For the input end of the DNN model, we study how the input features influence the model with the back-propagated  gradient signals ~\cite{jiewei2016visualizing}. The embedding output of the sparse feature ids (concatenated as $\mathbf{h}_{0}$) can be treated as the input for the following deep neural network. With the model fixed, for each input instance, we calculate the gradient of $\mathbf{h}_{0}$ with respective to the model's output $Pctr$:  
\begin{eqnarray}
\label{eqn:2.1}
\mathbf{g}_{0}= \nabla _{\mathbf{h}_{0}} Pctr.
\end{eqnarray}

\begin{figure*}[htb!]
\includegraphics[ width=6in]{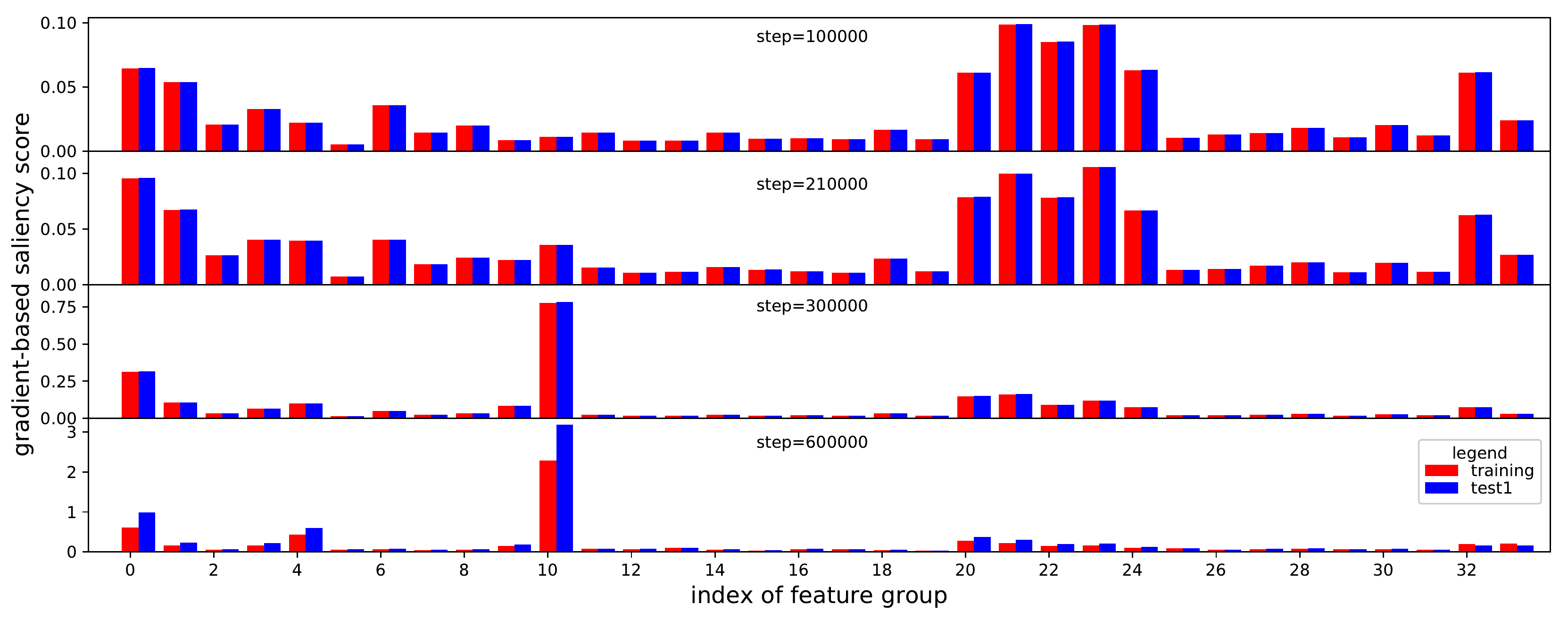}
\caption{Gradient-based saliency score of the $34$ feature groups for training and test1 sets.  Each bar represents a feature group. }
\label{fig:grademb}
\end{figure*}

The magnitude of each element of the gradient vector $\mathbf{g}_{0}$ quantifies the sensitivity of the model's output to the change in the particular embedding element. It describe how much a small change in a particular embedding value could affect the final output $Pctr$. Given a dataset, we calculate the saliency score for each feature group by averaging the mean absolute value of the corresponding $8$ gradient elements in $\mathbf{g}_{0}$ over the whole dataset. This saliency score provides us with an average measure of the model's sensitivity to each feature group for the given dataset.

We illustrate the saliency scores in Fig. \ref{fig:grademb}. Overall, the model is becoming increasingly sensitive to all the feature groups during training. In the overfitting regime, the score of feature group $10$ rises up dramatically and becomes much higher than the other feature groups. This feature group is composed of user ids, in which the number of ids is larger than any other feature group by at least two orders of magnitude ~\cite{ge2017image}. For this training stage, the model is trained to memorize the vast amount of information from user ids that is not generalizable, and thus significantly deteriorates the performance on test datasets.

\section{Discussion}
\label{sec:diss}

\subsection{Role of Layer $4$}
\label{sec:disslayer4}
The results about layer $4$ raise a question about the necessity to include this layer in the model. To answer this question, we modify the neural network and investigate the impact on performance of the retrained models. We modify layer $4$ by reducing or increasing its width by a factor of two, or even remove layer $4$ from the model. 
It turns out that these modifications do not affect the models' performance (highest test AUCs) for the different test dataset. Although not harmful, there is no benefit to include layer $4$ in the DNN model.

\subsection{Regularization}
\label{sec:dissreg}
Analysis in the previous section reveals that the model become over sensitive to the input when overfitting. Also, the high correlations among neurons for layer $3$ and $4$ (Fig. \ref{fig:corrstep}) imply that there might be severe co-adaptations ~\cite{srivastava2014dropout}. One may hope to use regularizations to control overfitting and obtain better performance on test data. We have tried L1 and  L2 regularization ~\cite{goodfellow2016deep}, and dropout ~\cite{srivastava2014dropout}, for a variety of hyper-parameters. However, no improvement is obtained. In future, more work needs to be conducted on improving model's generalization power.

\subsection{Feature Treatment}
\label{sec:dissfea}
Subsection \ref{sec:resgrad} discloses the problem that the model is greatly sensitive to the feature group of user ids when overfitting. Other than regularization, it is also possible to improve the models' generalization power by optimizing the input features. User id is a highly granular feature group. Inputting it directly to the embedding-based deep neural network may not be the optimal choice. Following the idea of Wide\&Deep ~\cite{widedeep}, we remove user id from the embedding layer. The bias of each user id is represented by a float number $b_{user}$ and added immediately into the output layer:

\begin{eqnarray}
\label{eqn:3.1}
Pctr = Sigmoid( \mathbf{W}_{5}\mathbf{h}_{4}+ b_{5}+b_{user}).
\end{eqnarray} 
This bias is trained jointly with the other parts of the model. We find this approach can improve AUC on the test datasets by about  $0.1\%$. 

\section{Applications}
\label{sec:app}
With the visualization and analysis techniques presented above, we discuss some of the practical applications in this section. 

\begin{itemize}[leftmargin=*]

\item The distribution of the predicted CTR score is very important for real-time bidding auctions. Understanding the score distribution can help us to design better 
calibration methods ~\cite{he2014practical,mcmahan2013ad}. Also, score distribution can help to find outliers or bad-fitted samples, which can in turn be used to improve the model.

\item Inspections of model's inner status and gradient signals open up the "black box" of the DNN model, helping us to understand the mechanism of the model and the influence of features. These approaches can be used to diagnose the model, like (but not limited to) underfitting/overfitting, gradient vanishing/explosion, ineffective model structure, etc..  A deep understanding of the model's mechanism can help us to design better model structure, training algorithm and features.

\item For online advertisting, it is of great importance to monitor the model's online performance and the health of data pipeline. Feeding the model with problematic data can cause disaster. However, it is very difficult to describe and monitor the distribution of the extremely sparse and high-dimensional data. Moreover, monitoring the model's online performance may not be sufficient. The model predicts CTR for hundreds of candidate ads for each biding, while only very few ads can win the bidding and get feedback from impression. The classic performance metrics are mainly based on those feedbacks, and thus can only cover a limited portion of biased data. 

The DNN model, by nature, transforms the sparse input data into dense numerical representations. Therefore, the statistics of neurons' output and the gradient signals can be implemented as a new kind of metrics to monitor the distribution of the input data. Note that no feedback labels are needed to calculate these quantities. For example, as illustrated in Fig. \ref{fig:stdlayer3}, the average standard deviation for layer $3$'s output changes with the naturally varying distribution of input data. Problematic input data can cause more significant change in  the statistics.

\end{itemize}

\section{Conclusion}
\label{sec:con}
In this work, we visualize and analyze a simple DNN model for CTR prediction down to neuron level. Model training and evaluations are performed over a series of datasets. The model is inspected from the output to the input end. The statuses of neurons are studied using a variety of methods. Gradients of the feature embeddings are used to create a salience map to describe the influence of the feature groups.  The analysis provides insightful knowledges of the model's mechanism, helping us to monitor, diagnose and refine the model. 

Currently, we are applying these approaches to build a model-based evaluation and monitoring system for our online advertising platform. Based on our industrial scenario, future work will focus on exploring more approaches to interpret deep learning, investigating more complex algorithms and applying these approaches to design better models and algorithms.

\bibliographystyle{ACM-Reference-Format}
\bibliography{qa-bibliography}


\begin{thebibliography}{29}


\ifx \showCODEN    \undefined \def \showCODEN     #1{\unskip}     \fi
\ifx \showDOI      \undefined \def \showDOI       #1{#1}\fi
\ifx \showISBNx    \undefined \def \showISBNx     #1{\unskip}     \fi
\ifx \showISBNxiii \undefined \def \showISBNxiii  #1{\unskip}     \fi
\ifx \showISSN     \undefined \def \showISSN      #1{\unskip}     \fi
\ifx \showLCCN     \undefined \def \showLCCN      #1{\unskip}     \fi
\ifx \shownote     \undefined \def \shownote      #1{#1}          \fi
\ifx \showarticletitle \undefined \def \showarticletitle #1{#1}   \fi
\ifx \showURL      \undefined \def \showURL       {\relax}        \fi
\providecommand\bibfield[2]{#2}
\providecommand\bibinfo[2]{#2}
\providecommand\natexlab[1]{#1}
\providecommand\showeprint[2][]{arXiv:#2}

\bibitem[\protect\citeauthoryear{Abadi, Agarwal, Barham, Brevdo, Chen, Citro,
  Corrado, Davis, Dean, Devin, et~al\mbox{.}}{Abadi et~al\mbox{.}}{2016}]%
        {abadi2016tensorflow}
\bibfield{author}{\bibinfo{person}{Mart{\'\i}n Abadi}, \bibinfo{person}{Ashish
  Agarwal}, \bibinfo{person}{Paul Barham}, \bibinfo{person}{Eugene Brevdo},
  \bibinfo{person}{Zhifeng Chen}, \bibinfo{person}{Craig Citro},
  \bibinfo{person}{Greg~S Corrado}, \bibinfo{person}{Andy Davis},
  \bibinfo{person}{Jeffrey Dean}, \bibinfo{person}{Matthieu Devin},
  {et~al\mbox{.}}} \bibinfo{year}{2016}\natexlab{}.
\newblock \showarticletitle{Tensorflow: Large-scale machine learning on
  heterogeneous distributed systems}.
\newblock \bibinfo{journal}{{\em arXiv preprint arXiv:1603.04467\/}}
  (\bibinfo{year}{2016}).
\newblock
\showURL{%
\url{https://www.tensorflow.org/}}


\bibitem[\protect\citeauthoryear{Alain and Bengio}{Alain and Bengio}{2016}]%
        {alain2016understanding}
\bibfield{author}{\bibinfo{person}{Guillaume Alain} {and}
  \bibinfo{person}{Yoshua Bengio}.} \bibinfo{year}{2016}\natexlab{}.
\newblock \showarticletitle{Understanding intermediate layers using linear
  classifier probes}.
\newblock \bibinfo{journal}{{\em arXiv preprint arXiv:1610.01644\/}}
  (\bibinfo{year}{2016}).
\newblock


\bibitem[\protect\citeauthoryear{Arras, Montavon, M{\"u}ller, and Samek}{Arras
  et~al\mbox{.}}{2017}]%
        {arras2017explaining}
\bibfield{author}{\bibinfo{person}{Leila Arras}, \bibinfo{person}{Gr{\'e}goire
  Montavon}, \bibinfo{person}{Klaus-Robert M{\"u}ller}, {and}
  \bibinfo{person}{Wojciech Samek}.} \bibinfo{year}{2017}\natexlab{}.
\newblock \showarticletitle{Explaining recurrent neural network predictions in
  sentiment analysis}.
\newblock \bibinfo{journal}{{\em arXiv preprint arXiv:1706.07206\/}}
  (\bibinfo{year}{2017}).
\newblock


\bibitem[\protect\citeauthoryear{Bahdanau, Cho, and Bengio}{Bahdanau
  et~al\mbox{.}}{2014}]%
        {bahdanau2014neural}
\bibfield{author}{\bibinfo{person}{Dzmitry Bahdanau},
  \bibinfo{person}{Kyunghyun Cho}, {and} \bibinfo{person}{Yoshua Bengio}.}
  \bibinfo{year}{2014}\natexlab{}.
\newblock \showarticletitle{Neural machine translation by jointly learning to
  align and translate}.
\newblock \bibinfo{journal}{{\em arXiv preprint arXiv:1409.0473\/}}
  (\bibinfo{year}{2014}).
\newblock


\bibitem[\protect\citeauthoryear{Bengio et~al\mbox{.}}{Bengio
  et~al\mbox{.}}{2009}]%
        {bengio2009learning}
\bibfield{author}{\bibinfo{person}{Yoshua Bengio} {et~al\mbox{.}}}
  \bibinfo{year}{2009}\natexlab{}.
\newblock \showarticletitle{Learning deep architectures for AI}.
\newblock \bibinfo{journal}{{\em Foundations and trends{\textregistered} in
  Machine Learning\/}} \bibinfo{volume}{2}, \bibinfo{number}{1}
  (\bibinfo{year}{2009}), \bibinfo{pages}{1--127}.
\newblock


\bibitem[\protect\citeauthoryear{Cheng and Koc}{Cheng and Koc}{2016}]%
        {widedeep}
\bibfield{author}{\bibinfo{person}{Heng-Tze Cheng} {and}
  \bibinfo{person}{Levent Koc}.} \bibinfo{year}{2016}\natexlab{}.
\newblock \showarticletitle{Wide \& deep learning for recommender systems}. In
  \bibinfo{booktitle}{{\em Proceedings of the ACM 1st Workshop on Deep Learning
  for Recommender Systems}}. \bibinfo{pages}{7--10}.
\newblock


\bibitem[\protect\citeauthoryear{Covington, Adams, and Sargin}{Covington
  et~al\mbox{.}}{2016}]%
        {youtube}
\bibfield{author}{\bibinfo{person}{Paul Covington}, \bibinfo{person}{Jay
  Adams}, {and} \bibinfo{person}{Emre Sargin}.}
  \bibinfo{year}{2016}\natexlab{}.
\newblock \showarticletitle{Deep neural networks for youtube recommendations}.
  In \bibinfo{booktitle}{{\em Proceedings of ACM Conference on Recommender
  Systems}}. \bibinfo{pages}{191--198}.
\newblock


\bibitem[\protect\citeauthoryear{Duchi, Hazan, and Singer}{Duchi
  et~al\mbox{.}}{2011}]%
        {duchi2011adaptive}
\bibfield{author}{\bibinfo{person}{John Duchi}, \bibinfo{person}{Elad Hazan},
  {and} \bibinfo{person}{Yoram Singer}.} \bibinfo{year}{2011}\natexlab{}.
\newblock \showarticletitle{Adaptive subgradient methods for online learning
  and stochastic optimization}.
\newblock \bibinfo{journal}{{\em Journal of Machine Learning Research\/}}
  \bibinfo{volume}{12}, \bibinfo{number}{Jul} (\bibinfo{year}{2011}),
  \bibinfo{pages}{2121--2159}.
\newblock


\bibitem[\protect\citeauthoryear{Ge, Zhao, Zhou, Chen, Liu, Yi, Hu, Liu, Sun,
  Liu, et~al\mbox{.}}{Ge et~al\mbox{.}}{2017}]%
        {ge2017image}
\bibfield{author}{\bibinfo{person}{Tiezheng Ge}, \bibinfo{person}{Liqin Zhao},
  \bibinfo{person}{Guorui Zhou}, \bibinfo{person}{Keyu Chen},
  \bibinfo{person}{Shuying Liu}, \bibinfo{person}{Huiming Yi},
  \bibinfo{person}{Zelin Hu}, \bibinfo{person}{Bochao Liu},
  \bibinfo{person}{Peng Sun}, \bibinfo{person}{Haoyu Liu}, {et~al\mbox{.}}}
  \bibinfo{year}{2017}\natexlab{}.
\newblock \showarticletitle{Image Matters: Jointly Train Advertising CTR Model
  with Image Representation of Ad and User Behavior}.
\newblock \bibinfo{journal}{{\em arXiv preprint arXiv:1711.06505\/}}
  (\bibinfo{year}{2017}).
\newblock


\bibitem[\protect\citeauthoryear{Glorot and Bengio}{Glorot and Bengio}{2010}]%
        {Glorot2010Understanding}
\bibfield{author}{\bibinfo{person}{Xavier Glorot} {and} \bibinfo{person}{Yoshua
  Bengio}.} \bibinfo{year}{2010}\natexlab{}.
\newblock \showarticletitle{Understanding the difficulty of training deep
  feedforward neural networks}.
\newblock \bibinfo{journal}{{\em Journal of Machine Learning Research\/}}
  \bibinfo{volume}{9} (\bibinfo{year}{2010}), \bibinfo{pages}{249--256}.
\newblock


\bibitem[\protect\citeauthoryear{Goodfellow, Bengio, and Courville}{Goodfellow
  et~al\mbox{.}}{2016}]%
        {goodfellow2016deep}
\bibfield{author}{\bibinfo{person}{Ian Goodfellow}, \bibinfo{person}{Yoshua
  Bengio}, {and} \bibinfo{person}{Aaron Courville}.}
  \bibinfo{year}{2016}\natexlab{}.
\newblock \bibinfo{booktitle}{{\em Deep Learning}}.
\newblock \bibinfo{publisher}{MIT Press}.
\newblock


\bibitem[\protect\citeauthoryear{Graepel, Candela, Borchert, and
  Herbrich}{Graepel et~al\mbox{.}}{2010}]%
        {Graepel:2010:WBC:3104322.3104326}
\bibfield{author}{\bibinfo{person}{Thore Graepel}, \bibinfo{person}{Joaquin
  Qui\~{n}onero Candela}, \bibinfo{person}{Thomas Borchert}, {and}
  \bibinfo{person}{Ralf Herbrich}.} \bibinfo{year}{2010}\natexlab{}.
\newblock \showarticletitle{Web-scale Bayesian Click-through Rate Prediction
  for Sponsored Search Advertising in Microsoft's Bing Search Engine}. In
  \bibinfo{booktitle}{{\em Proceedings of the 27th International Conference on
  International Conference on Machine Learning}} {\em
  (\bibinfo{series}{ICML'10})}. \bibinfo{publisher}{Omnipress},
  \bibinfo{address}{USA}, \bibinfo{pages}{13--20}.
\newblock
\showISBNx{978-1-60558-907-7}


\bibitem[\protect\citeauthoryear{He, Pan, Jin, Xu, Liu, Xu, Shi, Atallah,
  Herbrich, Bowers, et~al\mbox{.}}{He et~al\mbox{.}}{2014}]%
        {he2014practical}
\bibfield{author}{\bibinfo{person}{Xinran He}, \bibinfo{person}{Junfeng Pan},
  \bibinfo{person}{Ou Jin}, \bibinfo{person}{Tianbing Xu}, \bibinfo{person}{Bo
  Liu}, \bibinfo{person}{Tao Xu}, \bibinfo{person}{Yanxin Shi},
  \bibinfo{person}{Antoine Atallah}, \bibinfo{person}{Ralf Herbrich},
  \bibinfo{person}{Stuart Bowers}, {et~al\mbox{.}}}
  \bibinfo{year}{2014}\natexlab{}.
\newblock \showarticletitle{Practical lessons from predicting clicks on ads at
  facebook}. In \bibinfo{booktitle}{{\em Proceedings of the Eighth
  International Workshop on Data Mining for Online Advertising}}. ACM,
  \bibinfo{pages}{1--9}.
\newblock


\bibitem[\protect\citeauthoryear{Karpathy, Johnson, and Li}{Karpathy
  et~al\mbox{.}}{2015}]%
        {karpathy2015visualizing}
\bibfield{author}{\bibinfo{person}{Andrej Karpathy}, \bibinfo{person}{Justin
  Johnson}, {and} \bibinfo{person}{Fei-Fei Li}.}
  \bibinfo{year}{2015}\natexlab{}.
\newblock \showarticletitle{Visualizing and understanding recurrent networks}.
\newblock \bibinfo{journal}{{\em arXiv preprint arXiv:1506.02078\/}}
  (\bibinfo{year}{2015}).
\newblock


\bibitem[\protect\citeauthoryear{Koh and Liang}{Koh and Liang}{2017}]%
        {koh2017understanding}
\bibfield{author}{\bibinfo{person}{Pangwei Koh} {and} \bibinfo{person}{Percy
  Liang}.} \bibinfo{year}{2017}\natexlab{}.
\newblock \showarticletitle{Understanding Black-box Predictions via Influence
  Functions}. In \bibinfo{booktitle}{{\em International Conference on Machine
  Learning}}. \bibinfo{pages}{1885--1894}.
\newblock


\bibitem[\protect\citeauthoryear{Li, Chen, Hovy, and Jurafsky}{Li
  et~al\mbox{.}}{2016}]%
        {jiewei2016visualizing}
\bibfield{author}{\bibinfo{person}{Jiwei Li}, \bibinfo{person}{Xinlei Chen},
  \bibinfo{person}{Eduard Hovy}, {and} \bibinfo{person}{Dan Jurafsky}.}
  \bibinfo{year}{2016}\natexlab{}.
\newblock \showarticletitle{Visualizing and Understanding Neural Models in
  NLP}.
\newblock \bibinfo{journal}{{\em arXiv preprint arXiv:1506.01066v2\/}}
  (\bibinfo{year}{2016}).
\newblock


\bibitem[\protect\citeauthoryear{Maaten and Hinton}{Maaten and Hinton}{2008}]%
        {maaten2008visualizing}
\bibfield{author}{\bibinfo{person}{Laurens van~der Maaten} {and}
  \bibinfo{person}{Geoffrey Hinton}.} \bibinfo{year}{2008}\natexlab{}.
\newblock \showarticletitle{Visualizing data using t-SNE}.
\newblock \bibinfo{journal}{{\em Journal of machine learning research\/}}
  \bibinfo{volume}{9}, \bibinfo{number}{Nov} (\bibinfo{year}{2008}),
  \bibinfo{pages}{2579--2605}.
\newblock


\bibitem[\protect\citeauthoryear{Mahendran and Vedaldi}{Mahendran and
  Vedaldi}{2016}]%
        {mahendran2016visualizing}
\bibfield{author}{\bibinfo{person}{Aravindh Mahendran} {and}
  \bibinfo{person}{Andrea Vedaldi}.} \bibinfo{year}{2016}\natexlab{}.
\newblock \showarticletitle{Visualizing deep convolutional neural networks
  using natural pre-images}.
\newblock \bibinfo{journal}{{\em International Journal of Computer Vision\/}}
  \bibinfo{volume}{120}, \bibinfo{number}{3} (\bibinfo{year}{2016}),
  \bibinfo{pages}{233--255}.
\newblock


\bibitem[\protect\citeauthoryear{McMahan, Holt, Sculley, Young, Ebner, Grady,
  Nie, Phillips, Davydov, Golovin, et~al\mbox{.}}{McMahan
  et~al\mbox{.}}{2013}]%
        {mcmahan2013ad}
\bibfield{author}{\bibinfo{person}{Brendan McMahan}, \bibinfo{person}{Gary
  Holt}, \bibinfo{person}{David Sculley}, \bibinfo{person}{Michael Young},
  \bibinfo{person}{Dietmar Ebner}, \bibinfo{person}{Julian Grady},
  \bibinfo{person}{Lan Nie}, \bibinfo{person}{Todd Phillips},
  \bibinfo{person}{Eugene Davydov}, \bibinfo{person}{Daniel Golovin},
  {et~al\mbox{.}}} \bibinfo{year}{2013}\natexlab{}.
\newblock \showarticletitle{Ad click prediction: a view from the trenches}. In
  \bibinfo{booktitle}{{\em Proceedings of the 19th ACM SIGKDD international
  conference on Knowledge discovery and data mining}}. ACM,
  \bibinfo{pages}{1222--1230}.
\newblock


\bibitem[\protect\citeauthoryear{Pei, Cao, Yang, and Jana}{Pei
  et~al\mbox{.}}{2017}]%
        {pei2017deepxplore}
\bibfield{author}{\bibinfo{person}{Kexin Pei}, \bibinfo{person}{Yinzhi Cao},
  \bibinfo{person}{Junfeng Yang}, {and} \bibinfo{person}{Suman Jana}.}
  \bibinfo{year}{2017}\natexlab{}.
\newblock \showarticletitle{Deepxplore: Automated whitebox testing of deep
  learning systems}. In \bibinfo{booktitle}{{\em Proceedings of the 26th
  Symposium on Operating Systems Principles}}. ACM, \bibinfo{pages}{1--18}.
\newblock


\bibitem[\protect\citeauthoryear{Rauber, Fadel, Falcao, and Telea}{Rauber
  et~al\mbox{.}}{2017}]%
        {rauber2017visualizing}
\bibfield{author}{\bibinfo{person}{Paulo~E Rauber}, \bibinfo{person}{Samuel~G
  Fadel}, \bibinfo{person}{Alexandre~X Falcao}, {and}
  \bibinfo{person}{Alexandru~C Telea}.} \bibinfo{year}{2017}\natexlab{}.
\newblock \showarticletitle{Visualizing the hidden activity of artificial
  neural networks}.
\newblock \bibinfo{journal}{{\em IEEE transactions on visualization and
  computer graphics\/}} \bibinfo{volume}{23}, \bibinfo{number}{1}
  (\bibinfo{year}{2017}), \bibinfo{pages}{101--110}.
\newblock


\bibitem[\protect\citeauthoryear{Saxe, Bansal, Dapello, Advani, Kolchinsky,
  Tracey, and Cox}{Saxe et~al\mbox{.}}{2018}]%
        {michael2018on}
\bibfield{author}{\bibinfo{person}{Andrew~Michael Saxe},
  \bibinfo{person}{Yamini Bansal}, \bibinfo{person}{Joel Dapello},
  \bibinfo{person}{Madhu Advani}, \bibinfo{person}{Artemy Kolchinsky},
  \bibinfo{person}{Brendan~Daniel Tracey}, {and} \bibinfo{person}{David~Daniel
  Cox}.} \bibinfo{year}{2018}\natexlab{}.
\newblock \showarticletitle{On the Information Bottleneck Theory of Deep
  Learning}. In \bibinfo{booktitle}{{\em International Conference on Learning
  Representations}}.
\newblock
\showURL{%
\url{https://openreview.net/forum?id=ry_WPG-A-}}


\bibitem[\protect\citeauthoryear{Shan and Hoens}{Shan and Hoens}{2016}]%
        {deepcross}
\bibfield{author}{\bibinfo{person}{Ying Shan} {and} \bibinfo{person}{T~Ryan
  Hoens}.} \bibinfo{year}{2016}\natexlab{}.
\newblock \showarticletitle{Deep crossing: Web-scale modeling without manually
  crafted combinatorial features}. In \bibinfo{booktitle}{{\em Proceedings of
  ACM Conference on Knowledge Discovery and Data Mining}}.
\newblock


\bibitem[\protect\citeauthoryear{Shwartz-Ziv and Tishby}{Shwartz-Ziv and
  Tishby}{2017}]%
        {shwartz2017opening}
\bibfield{author}{\bibinfo{person}{Ravid Shwartz-Ziv} {and}
  \bibinfo{person}{Naftali Tishby}.} \bibinfo{year}{2017}\natexlab{}.
\newblock \showarticletitle{Opening the black box of deep neural networks via
  information}.
\newblock \bibinfo{journal}{{\em arXiv preprint arXiv:1703.00810\/}}
  (\bibinfo{year}{2017}).
\newblock


\bibitem[\protect\citeauthoryear{Srivastava, Hinton, Krizhevsky, Sutskever, and
  Salakhutdinov}{Srivastava et~al\mbox{.}}{2014}]%
        {srivastava2014dropout}
\bibfield{author}{\bibinfo{person}{Nitish Srivastava},
  \bibinfo{person}{Geoffrey Hinton}, \bibinfo{person}{Alex Krizhevsky},
  \bibinfo{person}{Ilya Sutskever}, {and} \bibinfo{person}{Ruslan
  Salakhutdinov}.} \bibinfo{year}{2014}\natexlab{}.
\newblock \showarticletitle{Dropout: A simple way to prevent neural networks
  from overfitting}.
\newblock \bibinfo{journal}{{\em The Journal of Machine Learning Research\/}}
  \bibinfo{volume}{15}, \bibinfo{number}{1} (\bibinfo{year}{2014}),
  \bibinfo{pages}{1929--1958}.
\newblock


\bibitem[\protect\citeauthoryear{Szegedy, Zaremba, Sutskever, Bruna, Erhan,
  Goodfellow, and Fergus}{Szegedy et~al\mbox{.}}{2013}]%
        {szegedy2013intriguing}
\bibfield{author}{\bibinfo{person}{Christian Szegedy},
  \bibinfo{person}{Wojciech Zaremba}, \bibinfo{person}{Ilya Sutskever},
  \bibinfo{person}{Joan Bruna}, \bibinfo{person}{Dumitru Erhan},
  \bibinfo{person}{Ian Goodfellow}, {and} \bibinfo{person}{Rob Fergus}.}
  \bibinfo{year}{2013}\natexlab{}.
\newblock \showarticletitle{Intriguing properties of neural networks}.
\newblock \bibinfo{journal}{{\em arXiv preprint arXiv:1312.6199\/}}
  (\bibinfo{year}{2013}).
\newblock


\bibitem[\protect\citeauthoryear{Tang, Shi, Wang, Feng, and Zhang}{Tang
  et~al\mbox{.}}{2017}]%
        {Tang2017Memory}
\bibfield{author}{\bibinfo{person}{Zhiyuan Tang}, \bibinfo{person}{Ying Shi},
  \bibinfo{person}{Dong Wang}, \bibinfo{person}{Yang Feng}, {and}
  \bibinfo{person}{Shiyue Zhang}.} \bibinfo{year}{2017}\natexlab{}.
\newblock \showarticletitle{Memory visualization for gated recurrent neural
  networks in speech recognition}.
\newblock \bibinfo{journal}{{\em Proceedings of IEEE International Conference
  on Acoustics, Speech and Signal Processing (ICASSP)\/}}
  (\bibinfo{year}{2017}).
\newblock


\bibitem[\protect\citeauthoryear{Tishby and Zaslavsky}{Tishby and
  Zaslavsky}{2015}]%
        {Tishby2015deep}
\bibfield{author}{\bibinfo{person}{Naftali Tishby} {and} \bibinfo{person}{Noga
  Zaslavsky}.} \bibinfo{year}{2015}\natexlab{}.
\newblock \showarticletitle{Deep learning and the information bottleneck
  principle}. In \bibinfo{booktitle}{{\em 2015 IEEE Information Theory Workshop
  (ITW)}}. \bibinfo{pages}{1--5}.
\newblock
\showDOI{%
\url{https://doi.org/10.1109/ITW.2015.7133169}}


\bibitem[\protect\citeauthoryear{Zeiler and Fergus}{Zeiler and Fergus}{2014}]%
        {zeiler2014visualizing}
\bibfield{author}{\bibinfo{person}{Matthew~D Zeiler} {and} \bibinfo{person}{Rob
  Fergus}.} \bibinfo{year}{2014}\natexlab{}.
\newblock \showarticletitle{Visualizing and understanding convolutional
  networks}. In \bibinfo{booktitle}{{\em European conference on computer
  vision}}. Springer, \bibinfo{pages}{818--833}.
\newblock


\end{thebibliography}

\end{document}